\documentclass[letterpaper, 10 pt, conference]{ieeeconf}  

\IEEEoverridecommandlockouts                              

\usepackage{graphicx}
\usepackage{booktabs}
\usepackage{caption}
\usepackage{multirow}
\usepackage {color}
\usepackage{amssymb}
\usepackage{amsmath}

\title{\LARGE \bf
 Hyperbolic Image-and-Pointcloud Contrastive Learning for 3D Classification
}

\author{Naiwen Hu$^\dagger$, Haozhe Cheng$^\dagger$, Yifan Xie, Pengcheng Shi  and Jihua Zhu$^{*}$
\thanks{$^\dagger$:equal contribute. $^{*}$:corresponding author(zhujh@xjtu.edu.cn). The authors are with School of 
Software Engineering, Xi'an Jiaotong University, Xi'an710000,
China and Shaanxi Joint Key Laboratory for Artifact Intelligence, China.}
}

\begin{document}

\maketitle
\thispagestyle{empty}
\pagestyle{empty}

\begin{abstract}

3D contrastive representation learning has exhibited remarkable efficacy across various downstream tasks. However, existing contrastive learning paradigms based on cosine similarity fail to deeply explore the potential intra-modal hierarchical and cross-modal semantic correlations about multi-modal data in Euclidean space. In response, we seek solutions in hyperbolic space and propose a hyperbolic image-and-pointcloud contrastive learning method (HyperIPC). For the intra-modal branch, we rely on the intrinsic geometric structure to explore the hyperbolic embedding representation of point cloud to capture invariant features. For the cross-modal branch, we leverage images to guide the point cloud in establishing strong semantic hierarchical correlations. Empirical experiments underscore the outstanding classification performance of HyperIPC. Notably, HyperIPC enhances object classification results by 2.8\% and few-shot classification outcomes by 5.9\% on ScanObjectNN compared to the baseline. Furthermore, ablation studies and confirmatory testing validate the rationality of HyperIPC's parameter settings and the effectiveness of its submodules.


\end{abstract}

\section{INTRODUCTION}

With the popularity of Foundation Model, self-supervised representation learning has achieved great success in the fields of natural language processing (NLP) \cite{devlin2018bert},\cite{raffel2020exploring}, computer vision \cite{he2022masked}, video signals \cite{oord2018representation}, and multi-modality \cite{he2020momentum},\cite{alayrac2022flamingo}. These methods use extreme amounts of data in the pre-training stage to obtain powerful representations for downstream tasks. In the 3D vision field, data collection and annotation are time-consuming and labor-intensive compared to 2D vision and NLP. Considering the issues of data scarcity and imbalance, it is challenging to obtain high-quality representations using self-supervised representation learning methods with limited data.

\begin{figure}
    \centering
    \includegraphics[width=0.45\textwidth]{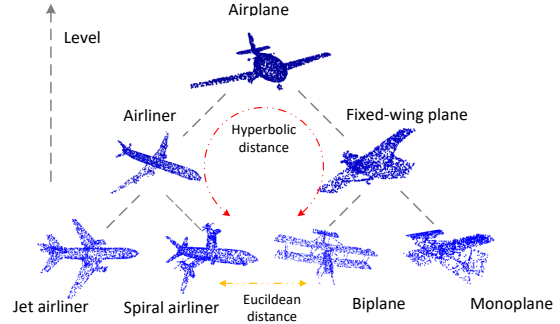}
    \vspace{-0.3cm}
    \caption{\textbf{Illustration of semantic hierarchy in hyperbolic space.} ``Airplane'' can be organized into a tree-like structure according to their flying modes and other semantic information, where the lower the level, the more detailed the object's description. The point cloud features located at different nodes of the same level should pass through the root node (red line) when calculating distance. However, the distance in Euclidean space is defined according to cosine similarity (yellow line).  }
    \label{fig:enter-label}
    \vspace{-0.7cm}
\end{figure}

Various 3D self-supervised representation learning methods have been developed. Contrastive learning encourages the representations of the same category to be closer and the representations of different categories to be farther apart \cite{chen2020simple},\cite{huang2021self}. Recently, many contrastive learning methods \cite{xie2020pointcontrast},\cite{chen20224dcontrast} have been proposed to deal with point cloud through various strategies for constructing positive-negative sample pairs.

It is well known that an image conveys various semantic information. Even for the same category of objects, humans can reason about their relative details and organize these concepts into a meaningful visual semantic hierarchy. As shown in Fig. \ref{fig:enter-label}, for the category of ``Airplane'', the point cloud located at a higher level has a more abstract description.

Unfortunately, current 3D self-supervised representation learning methods embed the point cloud in the Euclidean space using the same distance metric, which cannot capture the semantic hierarchy of the data. This may cause potential issues, as illustrated in Fig. \ref{fig:enter-label}, the specific concept (``Biplane'') is closer to other specific concepts (``Spiral airliners'') rather than the generic concept (``Airplane''). As a space with a constant negative curvature, the volume of hyperbolic space grows exponentially concerning the radius. Thus, the hyperbolic space can embed tree-like graphs with minimal distortion. To mine the latent semantic hierarchy in the point cloud, this property of hyperbolic space motivates us to embed point cloud representations into hyperbolic space.
 
In this work, we propose a simple and effective Hyperbolic Image-and-Pointcloud Contrastive Learning (HyperIPC) model. By projecting latent vectors to the hyperbolic space, we can efficiently extract the intrinsic semantic hierarchy of unlabeled data. We first map point cloud features from Euclidean space to the hyperbolic space and use the distance defined in the hyperbolic space for contrastive learning. Then, we compute their parent node in hyperbolic space, which is closer to the origin and can be regarded as a more abstract representation of the two different views. To leverage the semantic hierarchy information inherent in images, we employ a pre-trained image encoder to extract 2D information from rendered images, then map vectors to hyperbolic space for contrastive learning with the parent nodes. Moreover, to ensure the comprehensive exploitation of hyperbolic space, we optimize the nodes in Poincaré disk according to their level information.
 
The main contributions of our approach can be summarized as follows:
\begin{itemize}

    \item We propose HyperIPC, a simple and effective hyperbolic contrastive learning framework for self-supervised 3D point cloud pre-training incluing the intra-modal and cross-modal hyperbolic contrastive learning.

    \item We introduce the 2D VIT pre-trained by CLIP, which leverages the 2D knowledge to guide the point cloud to construct hierarchy in hyperbolic space.
    
    \item HyperIPC achieves state-of-the-art performance for contrastive learning on various downstream tasks, which indicates contrastive learning with hyperbolic distance outperforms the contrastive learning methods in the Euclidean space.

\end{itemize}

\section{RELATED WORK}

\begin{figure*}
    \centering
    \includegraphics[width=0.99 \linewidth]{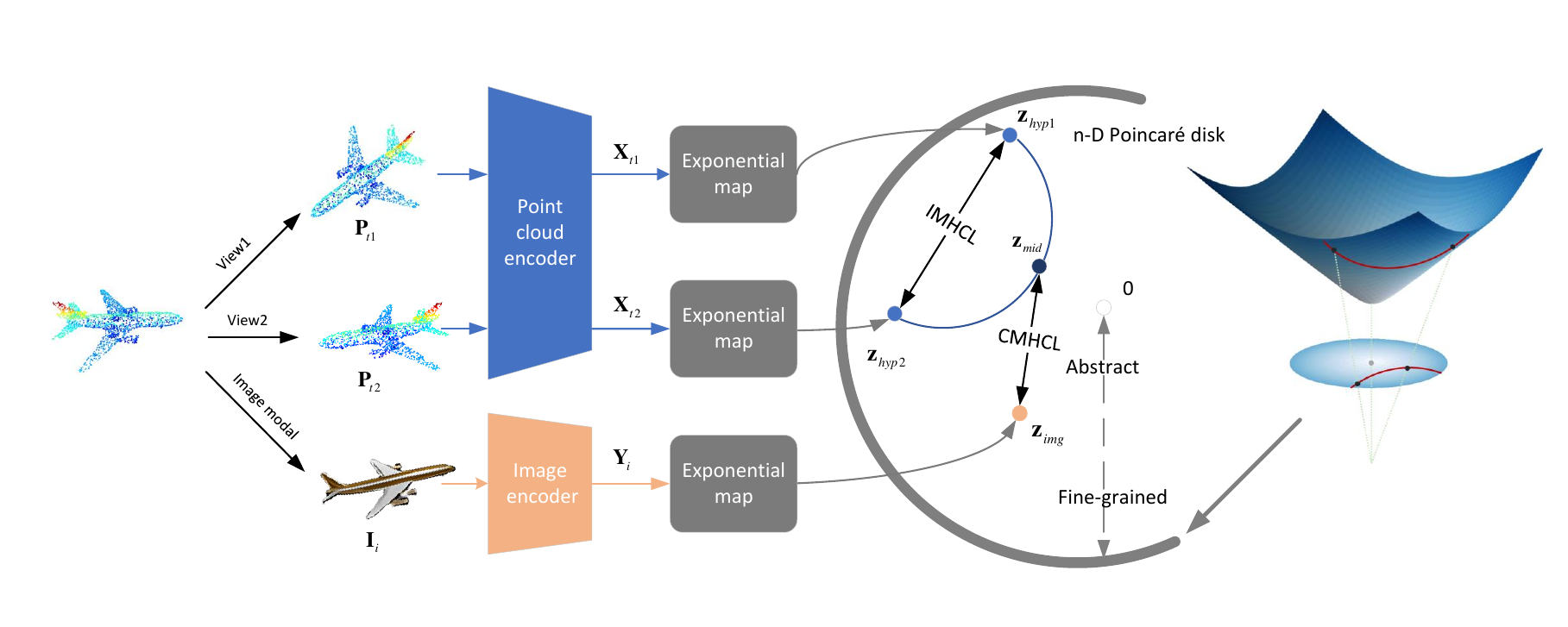}
    \vspace{-0.5cm}
    \caption{\textbf{Our proposed model architecture.} Point cloud branch: Intra-Modal Hyperbolic Contrastive Learning (IMHCL) makes the modal learn the invariance between two augmented point cloud. Image branch: Cross-Modal Hyperbolic Contrastive Learning (CMHCL) leverages rendered image guide point cloud to establish a hierarchical structure. }
    \label{fig:overall_architecture}
    \vspace{-0.5cm}
\end{figure*}

\subsection{Contrastive Learning in Point Cloud}

Contrastive learning leverages optimized contrastive loss to encourage augmentation of the same input to produce more comparable representations. In 3D vision, there are predominantly two categories of contrastive learning methods: object-level and scene-level. The former captures the global representation of the point cloud by treating the whole point cloud as an object \cite{lu2023joint},\cite{cheng2024multi}. For example, Du et al.\cite{du2021self} use self-similar patches within a single point cloud to facilitate semantic understanding. The latter focuses more on the interaction between point cloud and its scene \cite{xie2020pointcontrast},\cite{hou2021exploring}. In contrast to previous 3D contrastive learning methods, our HyperIPC extends the contrastive loss to the hyperbolic space.

\subsection{Hyperbolic Embedding}

With the development of deep learning, Euclidean space has become the standard manifold \cite{suris2021learning}. At the same time, hyperbolic space has been successfully applied to NLP \cite{tifrea2018poincar} due to the inherently hierarchical nature of language. HCNN \cite{chami2019hyperbolic} further extend deep neural network modules in the hyperbolic space. As a result, hyperbolic space has achieved success in image representation \cite{khrulkov2020hyperbolic},\cite{ermolov2022hyperbolic}. EDGCNet \cite{cheng2024edgcnet} proposes a dynamic hyperbric graph convolution module for 3D point cloud segmentation. Chen et al.\cite{chen2023self} propose a self-supervised learning method based on hyperbolic homotopy embedding to explore the nonlinear relationship of behavior trajectories. HIE \cite{yang2023hyperbolic} leverages the distance between data nodes and the origin node in hyperbolic space, deriving hierarchical information to optimize the existing hyperbolic models. In the 3D vision, HyCoRe \cite{montanaro2022rethinking} captures 3D part-whole hierarchy in supervised learning. In this paper, our HyperIPC aims to capture the semantic hierarchy among 3D objects in self-supervised representation learning.

\section{METHOD}

The schematic overview of the proposed method is depicted in Fig. \ref{fig:overall_architecture}. In this section, we first introduce how to obtain the hierarchical structure in the hyperbolic space. Then, we will discuss our overall framework diagram and describe the loss functions.

\subsection{Hyperbolic Geometric Embedding} 

We introduce hyperbolic space as the latent space to extract the latent semantic hierarchical information of the point cloud. In contrast to the Euclidean space with zero curvature, the n-dimensional hyperbolic space $H^{n}$ is a Riemannian manifold with constant negative curvature. The Poincaré, Lorentz, and Klein models are commonly used, equivalent representations of hyperbolic space, which can be interconverted and are suitable for different tasks \cite{cannon1997hyperbolic}. We use the Poincaré disk model $\left(\mathbb{D}_{c}^{n}, g^{\mathbb{D}}\right)$ because it can maintain numerical stability in the gradient-based learning process. The manifold $\mathbb{D}_{c}^{n}=\left\{x \in \mathbb{R}^{n}: c\|x\|^{2}<1, c \geq 0\right\}$ is equipped with a Riemannian metric $g^{\mathbb{D}}=\lambda_{c}^{x} g^{E}$, where $g^{E}$ is the metric tensor, $\lambda_{c}^{x}=\frac{2}{1-c\|x\|^{2}}$ is the conformal factor depending on the curvature $c$ and the position of the calculated point $x$ in the Poincaré disk. The metric of the points closer to the edge of the disk is scaled more by the conformal factor.

Conventional data operations are not applicable to the hyperbolic space, so we need to expand the operations in the hyperbolic space. Let $\|.\|$ be the Euclidean norm and $\langle.,.\rangle$ represent the Minkowski inner product. Given a pair of $\mathbf{x}, \mathbf{y} \in \mathbb{D}_{c}^{n}$, the addition operation is defined as:
\begin{equation}
\label{eq:add()}
    \mathbf{x} \oplus_{c} \mathbf{y}=\frac{\left(1+2 c\langle\mathbf{x}, \mathbf{y}\rangle+c\|\mathbf{y}\|^{2}\right) \mathbf{x}+\left(1-c\|\mathbf{x}\|^{2}\right) \mathbf{y}}{1+2 c\langle\mathbf{x}, \mathbf{y}\rangle+c^{2}\|\mathbf{x}\|^{2}\|\mathbf{y}\|^{2}}.
\end{equation}

The distance between $\mathbf{x}, \mathbf{y} \in \mathbb{D}_{c}^{n}$ in the hyperbolic space is defined as:
\begin{equation}
\label{eq:q()}
    D_{\text {hyp }}(\mathbf{x}, \mathbf{y})=\frac{2}{\sqrt{c}} 
    \operatorname{arctanh}\left(\sqrt{c}\left\|-\mathbf{x} \oplus_{c} 
    \mathbf{y}\right\|\right),
\end{equation}
in the above formulas, when the curvature $c$ approaches zero, the formulas for addition and distance become identical to those in the traditional Euclidean space. 

In the hyperbolic space, the geodesic is a generalization of the shortest path between two points or planes. As the curvature decreases, the distance between two points increases, and the geodesic is closer to the boundary \cite{lee2012smooth}. The midpoint of the geodesic between two points in hyperbolic space tends to be closer to the origin, resembling the concept of the tree \cite{suris2021learning},\cite{achlioptas2018learning}. This midpoint can be considered a more abstract parent node. This unique feature in the hyperbolic space can be reflected in Fig. \ref{fig:overall_architecture}. The point closer to the edge represents more specific categories, while closer to the origin represents more inductive instances. Given $n$ hyperbolic node vectors $\left(\mathbf{z}_{1}, \cdots, \mathbf{z}_{n}\right)$, the midpoint is computed in gyrovector space, which is given by:

\begin{equation}
\label{eq:center()}
    \mathbf{z}_{mid}=\frac{1}{2} \oplus_{c}\left(\frac{\sum_{i=1}^{n} \lambda_{c}^{x} \mathbf{z}_{i}}{\sum_{i=1}^{n} \left(\lambda_{c}^{x}-1\right)}\right).
\end{equation}

To map the embeddings to the hyperbolic space, we need to define a mapping relation from the Euclidean space to the Poincaré disk called the exponential map. The hyperbolic manifold $\mathbb{D}_{c}^{n}$ at $x$ has first order linear approximation tangent space $\mathcal{T}_{\mathbf{x}} \mathbb{D}_{c}^{n}\cong \mathbb{R}^{n}$, where $\mathcal{T}_{\mathbf{x}} \mathbb{D}_{c}^{n}=\left\{\mathbf{v} \in \mathbb{R}^{d}:\langle\mathbf{v}, \mathbf{x}\rangle=0\right\}$. The exponential map is defined as:
\begin{equation}
\label{eq:l()}
    \exp _{\mathbf{x}}^{c}(\mathbf{v})=\mathbf{x} \oplus_{c}\left(\tanh \left(\sqrt{c} \frac{\lambda_{\mathbf{x}}^{c}\|\mathbf{v}\|}{2}\right) \frac{\mathbf{v}}{\sqrt{c}\|\mathbf{v}\|}\right).
\end{equation}

\subsection{Model Architecture}

Due to the advantages of the hyperbolic space, our model consists of two branches: Intra-Modal Hyperbolic Contrastive Learning (IMHCL) and Cross-Modal Hyperbolic Contrastive Learning (CMHCL). These two branches obtain the semantic hierarchical structure of the point cloud from intra-modal and cross-modal perspectives in the hyperbolic space. Given the sample data $\mathcal{D}=\left\{\left(\mathbf{P}_{i}, \mathbf{I}_{i}\right)\right\}_{i=1}^{|\mathcal{D}|}$, where $\mathbf{P}_{i}$ denotes the 3D point cloud and $\mathbf{I}_{i}$ denotes the corresponding 2D image rendered from a random viewpoint. To enhance the discriminative ability of the point cloud encoder, we first apply IMHCL. Specifically, for a given point cloud sample $\mathbf{P}_{i}$, we apply random sampling transformations such as rotation, scaling, and translation to form two augmented point cloud $\mathbf{P}_{t1}$ and $\mathbf{P}_{t2}$.  
We use a shared-weight point cloud encoder to extract global features $\mathbf{X}_{t1}$ and $\mathbf{X}_{t2}$. Instead of regularizing the output in the Euclidean space, we use the exponential map Eq.(\ref{eq:l()}) to map the features from the Euclidean space to the hyperbolic space to obtain $\mathbf{z}_{hyp1}$ and $\mathbf{z}_{hyp2}$. Contrastive learning methods in Euclidean space define distance using squared Euclidean distance or cosine similarity. In hyperbolic space, we use Eq.(\ref{eq:q()}) to define the distance for contrastive learning.  

To prevent confusion, we first ignore the domain identifier $\mathbf{z}_{hyp1}$ and $\mathbf{z}_{hyp2}$. Given a positive sample pair $(i, j)$ and its representation $(\mathbf{z}_{i},\mathbf{z}_{j})$ in hyperbolic space, we define the hyperbolic contrastive learning loss function as:
\begin{equation}
    l(\mathbf{z}_{i}, \mathbf{z}_{j})=-\log \frac{\exp \left(-D_{\text {hyp }}\left(\mathbf{z}_{i}, \mathbf{z}_{j}\right) / \tau\right)}{\sum_{k=1, k \neq i}^{N} \exp \left(-D_{\text {hyp }}\left(\mathbf{z}_{i}, \mathbf{z}_{k}\right) / \tau\right)},
\end{equation}
where $D_{\text {hyp }}$ is the distance calculated by hyperbolic space, $\tau$ is the temperature coefficient, and $N$ represents the number of samples in a batch. The loss is calculated by all the positive samples $(i, j)$ and $(j, i)$.

\begin{equation}
    \mathcal{L}(\mathbf{z}_{i}, \mathbf{z}_{j})=\frac{1}{2N} \sum_{i=1}^{N}[l(\mathbf{z}_{i}, \mathbf{z}_{j}) + l(\mathbf{z}_{j}, \mathbf{z}_{i})].
\end{equation}

After obtaining embeddings of the same sample in hyperbolic space, the mean of the two embeddings $\mathbf{z}_{mid}$ is calculated using Eq.(\ref{eq:center()}). This mean point locate at the midpoint of the geodesic line between $\mathbf{z}_{hyp1}$ and $\mathbf{z}_{hyp2}$ is closer to the origin. This property is crucial for constructing a semantic hierarchy in hyperbolic space, similar to a tree structure where the mean of two leaf nodes represents a more general parent node rather than another leaf node.

Ermolov et al.\cite{ermolov2022hyperbolic} demonstrated that images have hierarchical information in hyperbolic space. We introduce the auxiliary CMHCL to guide the point cloud to establish a semantic hierarchy in hyperbolic space. During the model initialization process, the embeddings obtained by the image encoder are inaccurate. To prevent inaccurate 2D features guiding point cloud from being incorrectly embedded in hyperbolic space, the 2D pre-trained model is used to initialize the image encoder. We first use the visual encoder to obtain the embedding $\mathbf{Y}_{i}$ for the 2D image $\mathbf{I}_{i}$ of point cloud $\mathbf{P}_{i}$. Then, we apply the exponential map to project this Euclidean latent code $\mathbf{Y}_{i}$ into hyperbolic space to obtain $\mathbf{z}_{img}$. The goal of CMHCL is to maximize the similarity of $\mathbf{z}_{mid}$ to corresponding $\mathbf{z}_{img}$. In summary, CMHCL captures the image-pointcloud hierarchy to improve model discrimination.

\subsection{Hyperbolic Embedding Optimization}
\begin{figure}
    \centering
    \includegraphics[width=0.5\textwidth]{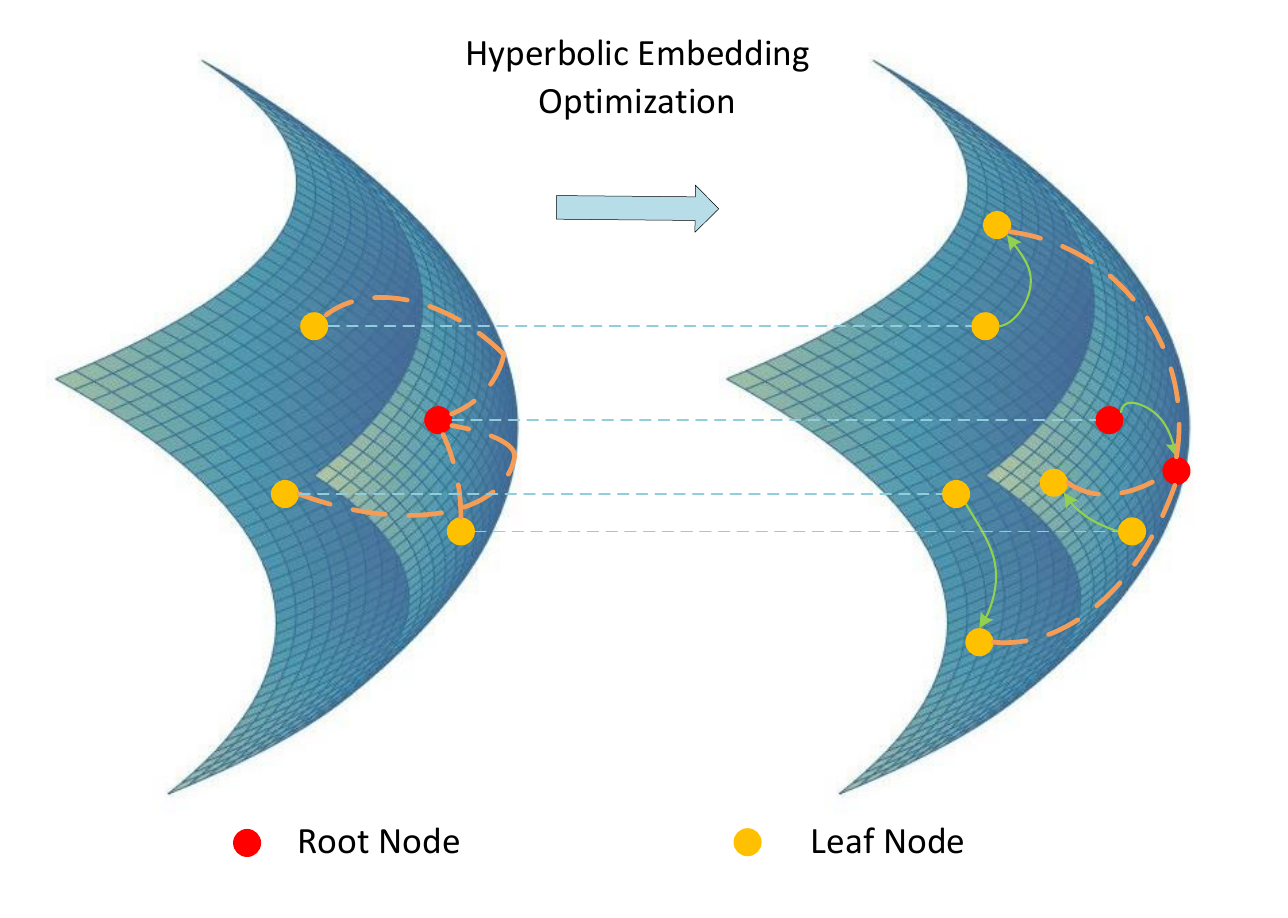}
    \vspace{-0.3cm}
    \caption{\textbf{Illustration of the hyperbolic embedding optimization.} Before optimization (Left), the root node deviates from the center of hyperbolic space and the leaf nodes are far from the boundary of the Poincaré disk. After hyperbolic optimization (Right), the root node of the data is aligned with the origin of the hyperbolic space, and the leaf nodes make full use of the characteristics of the hyperbolic space to disperse as much as possible.}
    \label{fig:icml}
    \vspace{-0.6cm}
\end{figure}

Since the hyperbolic space grows exponentially, the regions far from the origin are more spacious. The Leaf nodes in the tree structure occupy the majority and are as far from the origin as possible. Therefore, we hope that the point cloud embedding root node is optimized to the highest level, and the overall point cloud embedding should fully use the hyperbolic space's expansibility to disperse as much as possible. To solve the above problem, the first step is to identify the root node of the data, align it with the origin of the hyperbolic space, and then optimize the node according to their level information.

We first define the hyperbolic embedding center as the root node $\mathbf{z}_{c}$ by Eq.(\ref{eq:center()}). This node comes from the hyperbolic embedding and can be regarded as a super node connecting all subtrees. Then, we employ the root alignment strategy as defined:

\begin{equation}
    \overline{\mathbf{z}}=\mathbf{z}_{mid} \oplus_{c}\left(-\mathbf{z}_{c}\right).
\end{equation}

To efficiently access level information and guide hierarchical learning. We align the hyperbolic embedding center with the origin of the hyperbolic space, it reflects the relative distance between the leaf node and the root node, indicating its hierarchical level.

\begin{equation}
\label{eq:hdo()}
    \mathbf{z}_{hdo}=\frac{1}{|N|} \sum_{i \in N} w_{i} D_{\text {hyp }}(\overline{\mathbf{z}}_{i}, \mathbf{o}),
\end{equation} 
in Eq.(\ref{eq:hdo()}), $w_{i}$ indicates the node level in hyperbolic space which is computed by $\sigma(D_{\text {hyp }}(\overline{\mathbf{z}}_{i}, \mathbf{o}))$, the $\sigma$ is sigmoid function. The loss function is:

\begin{equation}
    \mathcal{L}_{\text {dho}}=\sigma\left(-\mathbf{z}_{hdo}\right).
\end{equation}

By optimizing the loss function, the high-level nodes close to the origin are assigned lower weights to prevent them from being pushed away. The low-level nodes far from the origin are assigned larger weights to help them reach correct positions in hyperbolic space.

\subsection{Overall Objective}

The overall loss function of the model consists of three parts: intra-modal hyperbolic contrastive loss, cross-modal hyperbolic contrastive loss, and hyperbolic embedding optimization loss, as follows:

\begin{equation}
\label{eq:obj()}
    \mathcal{L}=\mathcal{L}(\mathbf{z}_{hyp1}, \mathbf{z}_{hyp2})+\mathcal{L}(\mathbf{z}_{mid}, \mathbf{z}_{img})+\lambda \mathcal{L}_{\text {dho}},
\end{equation}
where $\lambda$ is a hyperparameter.

\vspace{0.4cm}
\section{Experiments}

\subsection{Pre-training}

\textbf{Dataset.} Our model is pretrained on ShapeNet\cite{yi2016scalable}, with over 50,000 CAD models in 55 categories. For given point cloud, a 2D image is randomly selected from the rendered images, captured from various viewpoint\cite{xu2019disn}. Each point cloud consists of 2,048 points with only $x, y, z$ coordinate, and the corresponding rendered image is resized to 224 × 224 pixels. Augmentation operations such as rotation and cropping are applied to increase the diversity of rendered images from random viewpoints.

\textbf{Implementation Details.} For a fair comparison with previous work, we apply DGCNN \cite{wang2019dynamic} as point cloud feature extractor, which exploits local geometric structures by constructing a local neighborhood graph and applying convolution-like operations on the edges connecting
neighboring pairs of points. As for the image encoder, we utilize VIT-S \cite{steiner2021train} that divides an image into patches, embeds them, and processes these embeddings through transformer layers to capture global image features. In addition, we use a two-layer MLP (384-128) as the projection head, and finally produce 128-dimensional feature projected in the hyperbolic space. The contrastive learning utilizes a curvature parameter $c = 0.1$, temperature $\tau = 0.2$ and the hyperbolic embedding optimization incorporates $\lambda = 0.01$.

For the image encoder, we use the AdamW optimizer \cite{loshchilov2017decoupled} with a learning rate value of $3\times10^{-5}$, a weight decay value of 0.01. We use the AdamW optimizer \cite{loshchilov2017decoupled} for the image encoder, with a learning rate of $1\times10^{-3}$ and a weight decay of $1\times10^{-4}$. All experiments are conducted on a single NVIDIA 3090Ti GPU with 100 epochs. After pretraining, we discard the image encoder and two projection heads. All downstream tasks are performed on the point cloud encoder.

\begin{table}[t]
\centering \small
\captionsetup{justification=justified}
\caption{\textbf{Comparison of ModelNet40 and ScanObjectNN linear classification results with previous self-supervised methods.} A linear classifier is fit onto the training split using the pretrained model and overall accuracy for classification in test split is reported. The dagger ($^{\dagger}$) denotes that the model was reproduced using DGCNN backbone.}
\label{table:table}
\begin{tabular}{l c c} 
 \toprule 
 Method & ModelNet40& ScanObjectNN\\
 \midrule
 3D-GAN\cite{wu2016learning} & 83.3 & - \\
 Latent-GAN\cite{achlioptas2018learning}  & 85.7 & - \\
 FoldingNet\cite{yang2018foldingnet}  & 88.4 & -\\
 DepthContrast\cite{zhang2021self}  & 85.4 & -\\
 ClusterNet\cite{zhang2019unsupervised}  & 86.8 & -\\
 STRL$^\dagger$\cite{huang2021spatio}  &  90.9 & 77.9 \\
 OcCo$^\dagger$\cite{wang2021unsupervised}  & 89.2 & 78.3\\
 CrossPoint$^\dagger$\cite{afham2022crosspoint}  & 91.2  &81.7\\
 CrossNet$^\dagger$\cite{wu2023self}  & 91.5 & 83.9\\
 \textbf{HyperIPC(Ours)$^\dagger$} & \textcolor{blue}{91.8}&\textcolor{blue}{84.5} \\
 \bottomrule
\end{tabular}
\vspace{-0.4cm}
\end{table}    
\vspace{-2mm}

\subsection{Downstream Tasks}

We evaluate the transferability of HyperIPC on two widely used downstream tasks in point cloud representation learning: (i) 3D object classification (synthetic and real-world), (ii) Few-shot object classification (synthetic and real-world)

\textbf{3D Object Classfication.} We demonstrate the generalizability of our approach in learning 3D shape representation from synthetic and real-world data through classification experiments on ModelNet40 \cite{wu20153d} and ScanObjectNN \cite{wu20153d}. ModelNet40 obtains point cloud by sampling 3D CAD models, and it contains 12,331 objects (9,843 for training and 2,468 for testing) from 40 categories. ScanObjectNN is more realistic and challenging for 3D point cloud classification, consisting of occluded objects from real-world indoor scans. It includes 2,880 objects (2,304 for training and 576 for testing) from 15 categories. 

\begin{figure*}
    \centering
    \includegraphics[width=0.99 \linewidth]{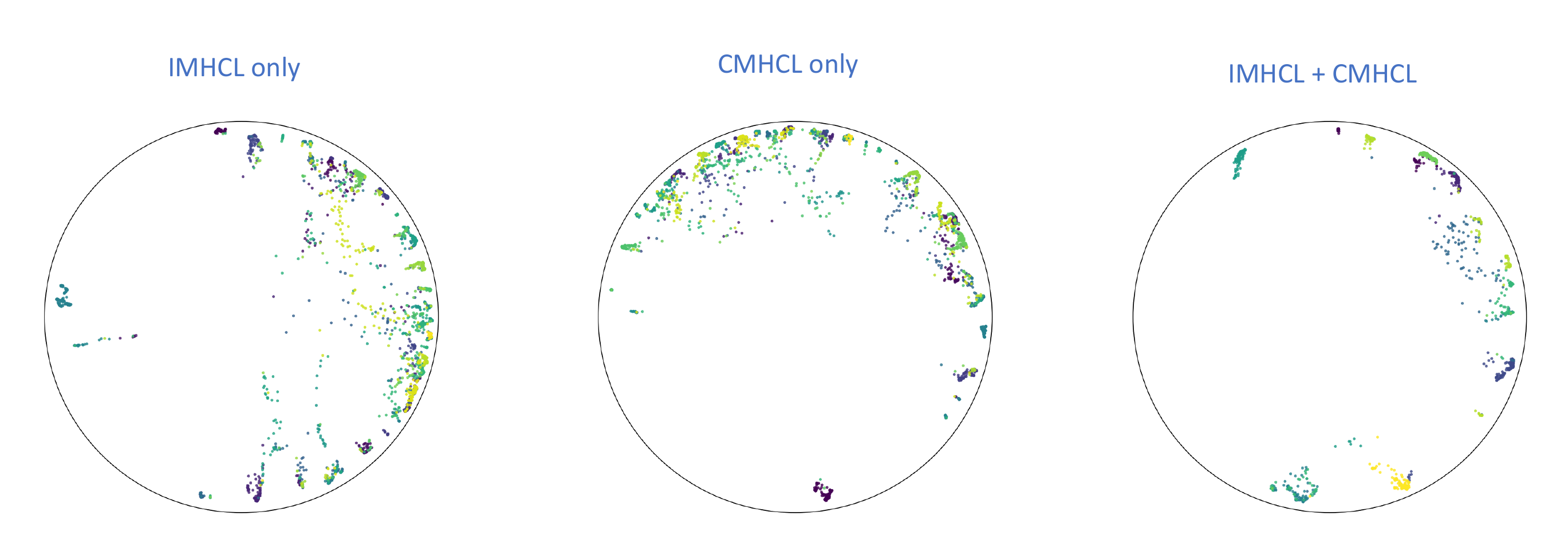}
    \caption{\textbf{UMAP\cite{mcinnes2018umap} embeddings for ModelNet10 (evaluation sets) on the Poincaré disk.} Each point inside the Poincaré disk corresponds to a sample. Different colors indicate different classes. After IMHCL and CMHCL, the samples are clustered according to the labels, and each category is also closer to the boundary of the Poincaré disk.}  
    \label{fig:vis}
    \vspace{-0.5cm}
\end{figure*}

We follow the standard protocols of STRL \cite{huang2021spatio}  and Crosspoint \cite{afham2022crosspoint} to test the accuracy of our network model in object classification. We freeze the point cloud encoder and fit the Support Vector Machine (SVM) classifier on the split of the training dataset. We randomly sample 1,024 points from each object for training and testing, with a batch size of 128 on the DGCNN backbone. Table. \ref{table:table} reports the linear classification results on ModelNet40 and ScanObjectNN. HyperIPC outperforms the previous state-of-the-art self-supervised methods in contrastive learning. More notably, we achieve 0.6$\%$ and 2.8$\%$ improvement over the baseline on the ModelNet40 and ScanObectNN. It can be observed that the underlying structure of real data tends to be hierarchical compared to synthetic datasets, hence leading to relatively more conspicuous results.

\textbf{Few-shot Object Classification.} We conduct Few-Shot Learning (FSL) experiments on the ModelNet40 and ScanObjectNN, using randomly selected $n$ classes from the dataset and $m$ samples from each class, with limited training data that can test the model's generalization ability. We perform ten FSL tasks and reported the mean and standard deviation for a fair comparison with previous methods \cite{sharma2020self},\cite{wang2021unsupervised}. Table. \ref{table:fewshot_cls} presents the FSL results on ModelNet40 and ScanObjectNN with the setting of $n \in\{5,10\}$ and $m \in\{10,20\}$. HyperIPC outperforms the previous work in few-shot classification tasks. These results demonstrate that HyperIPC can learn more discriminative latent representation with limited data, which can alleviate the overfitting issue and acquire semantic information of unknown data.
   


\begin{table}[t]
        \centering
        \caption{\textbf{Comparison of few-shot classification accuracy with existing methods on ModelNet40 and ScanObjectNN.} We performed ten few-shot tasks and report the mean and standard devia.}
        \begin{tabular}{l | c c | c c}
                \toprule
                \multirow{2}{*}{Method} & \multicolumn{2}{c|}{5-way} & \multicolumn{2}{c}{10-way} \\\cline{2-5}
                                        & 10-shot & 20-shot & 10-shot & 20-shot \\
                \midrule
                & \multicolumn{4}{c}{ModelNet40} \\
                Rand & 31.6$\pm$2.8 & 40.8$\pm$4.6 & 19.9$\pm$2.1 & 16.9$\pm$1.5\\
                Jigsaw\cite{sauder2019self} & 34.3$\pm$1.3 & 42.2$\pm$3.5 & 26.0$\pm$2.4 & 29.9$\pm$2.6\\
                cTree \cite{sharma2020self}  & 60.0$\pm$2.8 & 65.7$\pm$2.6 & 48.5$\pm$1.8 & 53.0$\pm$1.3\\
                OcCo\cite{wang2021unsupervised} & 90.6$\pm$2.8 & 92.5$\pm$1.9 & 82.9$\pm$1.3 & 86.5$\pm$2.2\\     
                CrossPoint\cite{afham2022crosspoint} & 91.0$\pm$2.9 & 95.0$\pm$3.4 & 82.2$\pm$6.5 & 87.8$\pm$3.0\\        
                ViPFormer\cite{10160658} & 91.1$\pm$7.2 & 93.4$\pm$4.5 & 80.8$\pm$4.2 & 87.1$\pm$5.8\\
                \textbf{HyperIPC(Ours)}  & \textcolor{blue}{94.5$\pm$4.4} & \textcolor{blue}{95.4$\pm$2.4} & \textcolor{blue}{86.7$\pm$4.6} & \textcolor{blue}{91.6$\pm$2.1}\\

                \midrule
                & \multicolumn{4}{c}{ScanObjectNN} \\
                Rand & 62.0$\pm$5.6 & 67.8$\pm$5.1 & 37.8$\pm$4.3 & 41.8$\pm$2.4\\
                Jigsaw\cite{sauder2019self} & 65.2$\pm$3.8 & 72.2$\pm$2.7 & 45.6$\pm$3.1 & 48.2$\pm$2.8\\
                cTree\cite{sharma2020self} & 68.4$\pm$3.4 & 71.6$\pm$2.9 & 42.4$\pm$2.7 & 43.0$\pm$3.0\\
                OcCo\cite{wang2021unsupervised} & 72.4$\pm$1.4 & 77.2$\pm$1.4 & 57.0$\pm$1.3 & 61.6$\pm$1.2\\     
                CrossPoint~\cite{afham2022crosspoint} & 72.5$\pm$8.3 & 79.0$\pm$1.2 & 59.4$\pm$4.0 & 67.8$\pm$4.4\\     
                ViPFormer\cite{10160658} & 74.2$\pm$7.0 & 82.2$\pm$4.9 & 63.5$\pm$3.8 & 70.9$\pm$3.7\\
                \textbf{HyperIPC(Ours)}  & \textcolor{blue}{79.6$\pm$7.6} & \textcolor{blue}{86.0$\pm$6.0} & \textcolor{blue}{68.0$\pm$4.6} & \textcolor{blue}{75.7$\pm$3.5}\\
                \bottomrule
        \end{tabular}
        \label{table:fewshot_cls}
        \vspace{-0.3cm}
\end{table}
\subsection{Ablations and Analysis}

In this section, we report the results of the ablation experiments. We use DGCNN as the point cloud feature extractor in all the classification experiments, with Linear SVM on ScanObjectNN.

\textbf{Impact of joint learning objective.}

\begin{figure}
    \centering
    \includegraphics[width=0.5\textwidth]{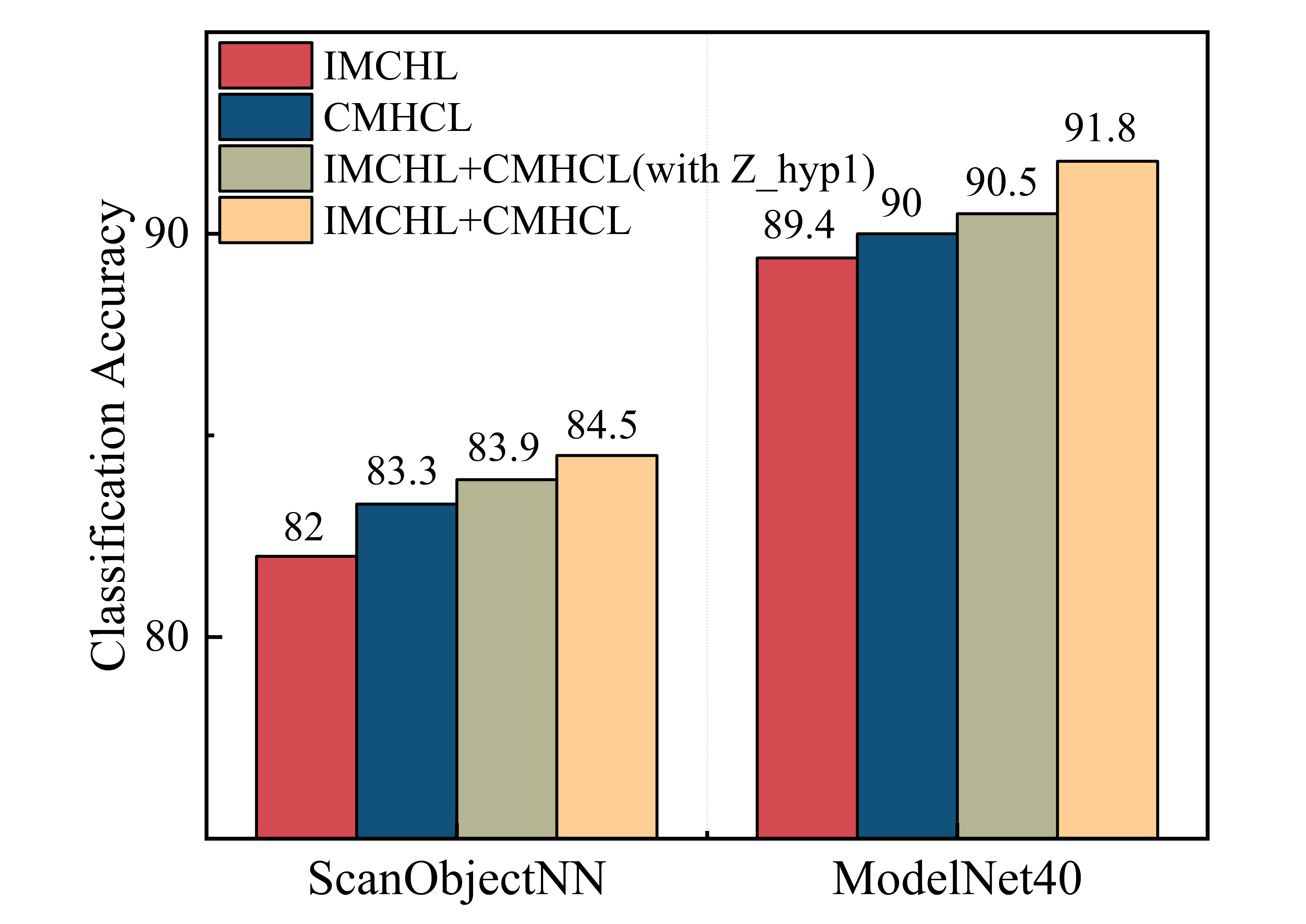}
    \vspace{-0.5cm}
    \caption{\textbf{Impact of joint learning objective.} Classification accuracy of intra-modal and cross-modal and joint learning objectives on ScanObjectNN and ModelNet40. }
    \label{fig:compare}
    \vspace{-0.8cm}
\end{figure}

We hypothesize that joint learning objectives in hyperbolic space exhibit a more discernible capacity than separate learning objectives. IMHCL encourages the model to acquire the invariance of the point cloud, making the distance between the point cloud of the same category close in hyperbolic space. CMHCL provides the point cloud encoder with semantic information from the visual domain, helping to establish a hierarchical structure in hyperbolic space. The joint learning in hyperbolic space can produce better results as shown in Fig.\ref{fig:compare}. To verify the CMHCL captures the semantic hierarchy, we compare the proposed objective Eq.(\ref{eq:obj()}) to the objective that replaces $\mathbf{z}_{mid}$ with $\mathbf{z}_{hyp1}$ in Eq.(\ref{eq:obj()}). It can be observed that the representations not only capture the semantic level of the data but also incorporate the information of the 2D images. Fig. \ref{fig:vis} illustrates how learned embeddings are arranged on the Poincaré disk. Compared with single hyperbolic contrastive learning method, the joint hyperbolic contrastive learning approach clusters samples according to their labels more effectively. Each category is also closer to the boundary of the Poincaré disk, indicating that the encoder has successfully separated the classes. 

\begin{table}[h]
    \centering
    \caption{\textbf{The experimental results under different curvatures.}} 
    \label{table:multiple_images}
    \vspace{-0.2cm}
    \begin{tabular}{cccccc}
    \toprule
    \begin{tabular}{l}$c$\end{tabular} & 0.01 & 0.1 & 0.3 & 0.5 & 1.0\\ \hline
    Accuracy & 84.0 & \textbf{84.3} & 83.8 & 83.7 & 83.7 \\ 
    \bottomrule
    \vspace{-0.4cm}
    \end{tabular}
\end{table}
\textbf{Curvatures.}

Table. \ref{table:multiple_images} shows the results of the model under different curvatures $c$. Our model is robust when the curvature is small, while larger values cause the model to degrade. It is worth noting that when the curvature is small and gradually approaches zero, the radius of the hyperbolic space becomes infinite and tends to the Euclidean space, which provides better stability.

\textbf{Image Encoder.}

In Crosspoint\cite{afham2022crosspoint}, initializing the image encoder with a normal distribution leads to inaccurate image embedding, resulting in an unsatisfactory hierarchical structure of the point cloud in hyperbolic space. We introduce the CLIP \cite{radford2021learning} pre-trained model as the image encoder, which can utilize the information implied by the images. CLIP \cite{radford2021learning} employs a two-tower network that aligns global representation of languages and images using extensive data. As shown in Table. \ref{table:detach}, updating the parameters of the image encoder during training allows for accurate semantic hierarchical information to be captured in hyperbolic space.

\section{Conclusion}

In this paper, we propose a simple yet effective method to capture the point cloud semantic hierarchy in hyperbolic space. After learning the semantic hierarchy from images, our model can continuously edit the semantic hierarchical features of the point cloud, achieving better results and more discriminating models. Experiments demonstrate that our model outperforms methods that use Euclidean representations. In future work, we will explore combining hyperbolic space with generative models and addressing segmentation tasks within hyperbolic space.

 \begin{table}[!t]
\caption{ \textbf{The experimental results under different image encoder.}} 
\label{table:detach}
\vspace{-0.1cm}
\begin{center}
\resizebox{0.72\linewidth}{!}{
\begin{tabular}{ccc}
\toprule[0.95pt]
Backbone & Back-propagation & Accuracy\\
\midrule[0.6pt]
VIT-S\cite{radford2021learning} & $\checkmark$ & \textbf{84.3}\\
VIT-S\cite{radford2021learning} & $\times$ & 83.7\\
Resnet\cite{he2016deep} & $\checkmark$ & 81.7\\
\bottomrule[0.95pt]
\end{tabular}
}
\end{center}
\vspace{-0.6cm}
\end{table}

\section{Acknowledgements}
This work was supported by the Key Research and Development Program of Shaanxi Province under Grants 2021GY-025. 





\bibliographystyle{IEEEtran}
\bibliography{IEEEabrv}

\end{document}